# Developing an aeroponic smart experimental greenhouse for controlling irrigation and plant disease detection using deep learning and IoT

**Mohammadreza Narimani[1], Ali Hajiahmad[1], Ali Moghimi[2,*], Reza Alimardani[1], Shahin Rafiee[1], Amir Hossein Mirzabe[1]**

[1]*Department of Mechanics of Biosystem Engineering, Faculty of Engineering & Technology, College of Agriculture & Natural Resources, University of Tehran, Karaj, Alborz, Iran*

[2]*Department of Biological and Agricultural Engineering, University of California, Davis, One Shields Ave, Davis, CA 95616 USA*

ABSTRACT.

*Controlling environmental conditions and monitoring plant status in greenhouses is critical to promptly making appropriate management decisions aimed at promoting crop production. The primary objective of this research study was to develop and test a smart aeroponic greenhouse on an experimental scale where the status of Geranium plant and environmental conditions are continuously monitored through the integration of the internet of things (IoT) and artificial intelligence (AI). An IoT-based platform was developed to control the environmental conditions of plants more efficiently and provide insights to users to make informed management decisions. In addition, we developed an AI-based disease detection framework using VGG-19, InceptionResNetV2, and InceptionV3 algorithms to analyze the images captured periodically after an intentional inoculation. The performance of the AI framework was compared with an expert's evaluation of disease status. Preliminary results showed that the IoT system implemented in the greenhouse environment is able to publish data such as temperature, humidity, water flow, and volume of charge tanks online continuously to users and adjust the controlled parameters to provide an optimal growth environment for the plants. Furthermore, the results of the AI framework demonstrate that the VGG-19 algorithm was able to identify drought stress and rust leaves from healthy leaves with the highest accuracy, 92% among the other algorithms.*

# 1. Introduction

In order to meet the increasing demand for agricultural products, critical challenges in crop production must be addressed, such as shortage of water and land resources, controlling plant diseases, and reducing the use of fertilizers and chemicals. In recent decades, soilless cultivation methods have been considered by scientists and growers to overcome water and land scarcity (Michelon et al., 2020; Moncada et al., 2020; Xie et al., 2020) mainly because these methods do not require fertile soil; these systems require less space and water than conventional methods. The lower costs and the possibility of vertical cultivation are other advantages of these systems (Barbosa et al., 2015; Zimmermann & Fischer, 2020). The possibility of vertical cultivation can increase the potential of their construction near urban areas and reduce the cost of transporting crops to urban areas.

The aeroponic system is one of the soilless cultivation methods in controlled growing environments, which is the most modern cultivation method of its kind (Jamshidi et al., 2020; Koukounaras, 2021; Lakhiar et al., 2020). There is no solid phase in the aeroponic method, and the roots grow in a biphasic environment (liquid and air). Constant contact with oxygen triggers metabolic processes that positively affect root growth and nutrient uptake (Komosa et al., 2014). Different beneficial effects of using the aeroponic method for various plants have been reported. Results of aeroponic application for tomatoes, lettuce, cucumber, and strawberry (Giacomelli & Smith, 1989; Massantini, 1985; Repetto et al., 1993), leafy vegetables (Demšar et al., 2004; Lim, 1996), potato (Farran & Mingo-Castel, 2006; Ritter et al., 2001), Medicinal plants (Christie & Nichols, 2003; Hayden, 2006), Ornamental plants (Christie & Nichols, 2003; Fascella & Zizzo, 2006; Kreij & Hoeven, 1997), tree and shrub species (Barak et al., 1996; Martin-Laurent et al., 2000; Soffer & Burger, 1988) have been reported to be successful.

The enormous benefits of aeroponic systems include higher yield, independence from soil and farm, water productivity, the possibility of production at any time of the year and independence from environmental conditions, possibility of examining root yield during biological research, reduction of root environment temperature in tropical regions, and saving on fertilizer use and chemical inputs due to recirculation of nutrient solution in the system (Cao et al., 2020; Lakhiar et al., 2020; Partap et al., 2020). According to previous studies, aeroponic systems can save 98% of water, 60% of nutrients, and 100% pesticides and herbicides (Spinoff, 2006).

Although an aeroponic system alone offers several advantages over conventional cultivation methods, increasing productivity in these systems and achieving the optimum point depends on optimal monitoring of environmental conditions on an instantaneous basis. In order to achieve optimal plant growth conditions and maximum production of plant products, accurate knowledge of all environmental factors affecting plant yield and growth (temperature, humidity, oxygen, carbon dioxide, light, nutrients in the irrigation solution, and disease) is essential. Monitoring all environmental conditions, analyzing the relevant data, and making the necessary changes in accordance with the conditions of optimal plant growth by human resources is very time-consuming and costly. These limitations have resulted in controlling only a few environmental factors in most greenhouses, which is a significant obstacle to achieving economic production and maximum efficiency. Also, the parameters are monitored in a coarse temporal resolution, which reduces the probability of a proportionate reaction at the appropriate time and increases the economic risk. Another limitation of conventional greenhouses is the need to monitor the condition of the greenhouse on-site, which increases the labor costs, can delay the response time, and is prone to human error. In order to mitigate these limitations, all environmental factors in the greenhouse can be remotely and autonomously monitored on an instantaneous basis using the internet of things (IoT) (Raviteja & Supriya, 2020; Rayhana et al., 2020; S. Wang et al., 2020).

The main limitation of instantaneous monitoring of the large datasets streamed on the cloud is the complexity and interpretation of the data. In this regard, the use of new methods such as artificial intelligence and data mining (Lu et al., 2019; Tik et al., 2009; Zaini et al., 2020; Afonso et al., 2020; Hamrani et al., 2020; L. Wang et al., 2020), has been considered by researchers. For instance, one of the most effective ways to identify plant diseases in a controlled environment is to implement AI techniques to facilitate the analysis process of large datasets captured by various sensing technologies. The results of the previous studies show that the use of new image processing methods using machine learning algorithms and deep learning offers the ability to diagnose plant diseases immediately after the emergence of diseases (Chethan et al., 2021; Gargade & Khandekar, 2021; Pineda et al., 2021; Srinivas et al., 2021).

The diagnosis of plant diseases in the early stages of growth is one of the most critical parameters of the culture medium. Many studies have been performed to diagnose plant diseases using machine learning and deep learning techniques. In one of the studies (Aversano et al., 2020), the general PlantVillage database was used, including nine groups of diseased tomato leaves and one group of healthy tomato leaves. In this study, three pre-trained convolutional neural networks VGG-19, Xception, and ResNet-50, were used to demonstrate the power of transference learning. The results showed that the two models Xception and VGG-19, achieved very high accuracy of 0.95 and 0.97, respectively. In contrast, the performance accuracy of the ResNet-50 algorithm was not optimal and was approximately 0.6.

The main objectives of this study were to (i) automate the process of monitoring environmental conditions and maintain optimal plant growth conditions using IoT in an aeroponic innovative experimental greenhouse for Geranium production, and (ii) develop an AI-based framework to identify rust disease and drought stress in Geranium leaves.

## 2. Method and materials

### 2.1 Design and implementation of smart greenhouse

In order to provide an environment for plant growth, indoor structures with an area of 9 square meters with dimensions of 3x3 square meters and 2 meters high were considered in the campus of agriculture and natural resources of the University of Tehran. The use of black polycarbonate sheets was recommended to prevent the entry of natural or artificial light into the greenhouse to create the ability to control the conditions of artificial light. Light sources with a distance of 1 meter from each other were installed in 9 points of the greenhouse roof in a row that can provide 8000 lux of light.

Temperature is one of the most critical parameters in the greenhouse. An electric heating system based on electric resistance was used to heat the greenhouse space. Also, a cooling system based on cellulose ventilation is considered. According to the information received from the sensors, the cooling and heating systems installed in the greenhouse receive their active and inactive command from the central processing unit. Humidity is another important environmental factor that in this project, ultrasonic humidifier was used to provide humidity. These humidifiers are installed at the height of 150 cm in different parts of the greenhouse and receive their command to be active and inactive from the central processing system.

To control the temperature and humidity in the greenhouse environment, installing sensors in different greenhouse parts is necessary. In this regard, three temperature-humidity sensors were installed in different parts of the greenhouse at certain distances and heights from the plant. Besides, one electric fan was installed on the roof of the greenhouse for ventilation. The output information from all sensors is sent to the central processing unit, and after processing this information, it is decided whether the heating, cooling, humidity generation, and ventilation systems should be active or not. In addition, all commands issued by the central processing unit were notified to the user through the network and stored in the cloud space and external memory of the system to calculate the momentary energy consumption in different stages of plant growth.

The most important part of an aeroponic greenhouse is the location of the plant. Polymer cultivation boxes with dimensions of 53 cm in length, 33 cm in width and 28 cm in height were used for seating the plant and feed it by the aeroponic method.

To implement the aeroponic system based on the centrifugal method, several centrifugal nozzles were placed inside each box, and the nutrient solution was transferred to the nozzles by the necessary pressure, flow and speed by a centrifugal pump. Centrifugal pumps equipped with asynchronous electromotor with a power of 0.25 kW are responsible for transferring the nutrient solution to the nozzles. To provide the required amount of nutrient solution flow, it is necessary to consider an electric pump for each box. Centrifugal nozzles are mounted on a frame at the top of the box (close to the plant root location), and the nutrient solution drips out of the nozzles and raises the moisture at the root of the plant. A large number of droplets settle on the roots and are absorbed. Eventually, a large number of droplets accumulate on the bottom of the container due to gravity. To reuse the nutrient solution and prevent the formation of polluted water, the nutrient solution can be sucked by installing another electric pump in each box and transferred to the storage tanks through return line pipes. The schematic of an aeroponic feeding box based on the centrifugal method is shown in Figure 1.

Polycarbonate plates, storage baskets and polymer foams were used to place the plant in aeroponic boxes. Polymer plates with the dimension of 53 cm in length and 33 cm in width are cut and placed on culture boxes. Twelve holes were created on each page by a circular cutter. Inside each of these holes, a number of holding baskets is placed. The geometric shape of the holding basket is such that the upper edge has a larger diameter than the diameter of the hole created in the polycarbonate plates, and its wall has a smaller diameter than the holes created. This way, the top edge of the basket can rest on the plate. The walls and floor of the basket are reticulated and not only allow fog particles or soluble droplets to enter and exit, but the plant roots will be able to exit through its holes. The inner diameter of the baskets is larger than the outer diameter of the plant stem. In this case, polymer foams will be used to prevent the displacement and removal of plant degrees of freedom and the escape of fog droplets. The use of polymer foams during plant growth will not cause significant stress to the stems as the stem diameter increases. The typical and explosive view of an aeration box with centrifugal irrigation technique after plant placed is shown in Figure 1.

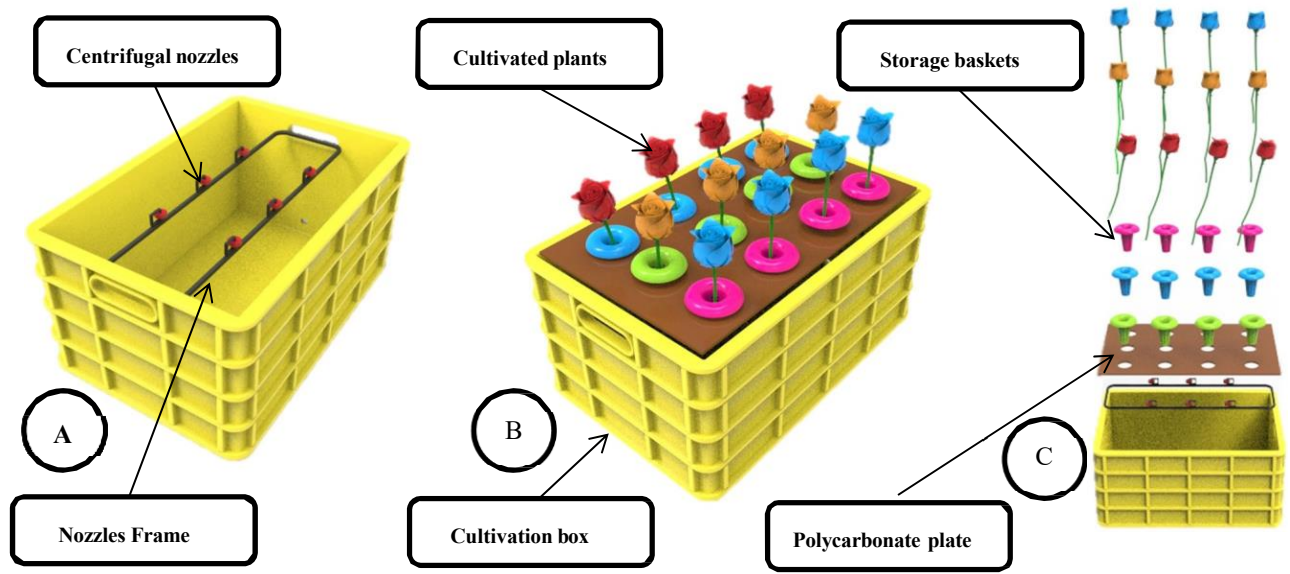

**Figure 1. Scheme of cultivation box equipped with centrifugal irrigation (A) cultivation box before planting geranium, (B) cultivation box after planting geranium, (C) explosive view of cultivation box**

In addition to the above, the use of ultraviolet light to eliminate a variety of contaminants is considered by researchers. Ultraviolet rays affect all microorganisms such as bacteria, yeasts, fungi, algae and even viruses and cause widespread disinfection. Therefore, it is necessary to place several ultraviolet lamps in storage tanks. The ultraviolet wavelength, intensity and duration of lighting are important factors to control. According to the above, the general view of the centrifugal irrigation part of the greenhouse is shown in Figure 2.

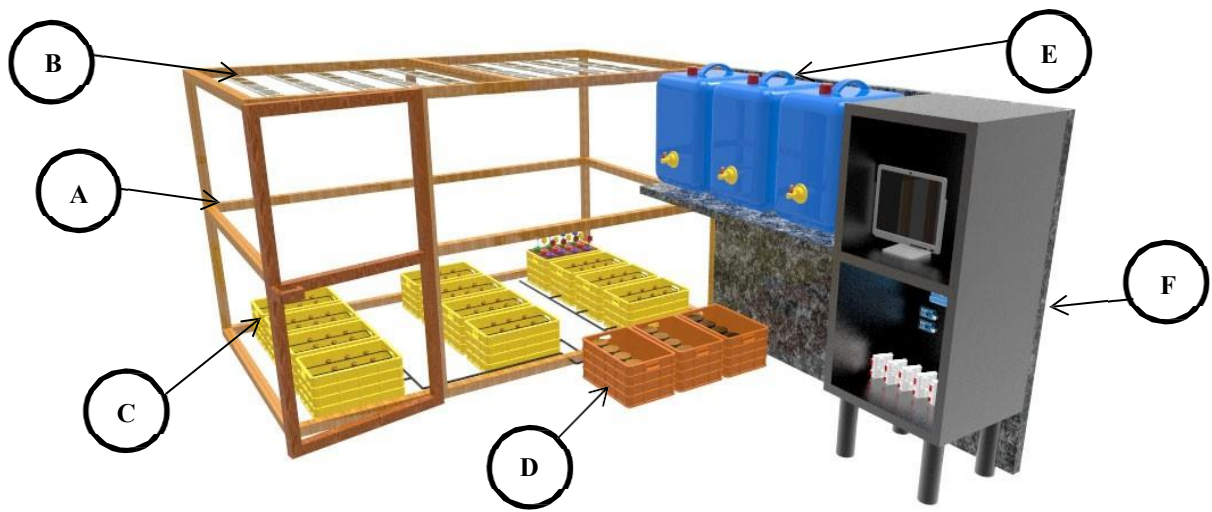

**Figure 2. Overview of the centrifugal irrigation part of greenhouse (A) Greenhouse chassis which three SHT75 sensors for monitoring temperature and humidity are installed on it, (B) LED strips for artificial lighting, (C) Cultivation boxes, (D) Nutrient solution storage boxes which three YF-S201 sensors for monitoring water flow are installed on each of them, (E) Nutrient solution storage tanks which three SRF05 sensors for monitoring water volume are installed on them (F) Information processing, storage and sending unit by IoT system**

## 2.2  Implementation of intelligent irrigation system

Cultivation with different irrigation schedules in the greenhouse is necessary to evaluate the functional performance of the aeroponic system and equipment used to monitor environmental conditions. In the aeroponic system, irrigation with on and off intervals of 10 minutes and 5 minutes is performed by asynchronous pumps and water transfer to centrifugal nozzles according to Figure 3, respectively.

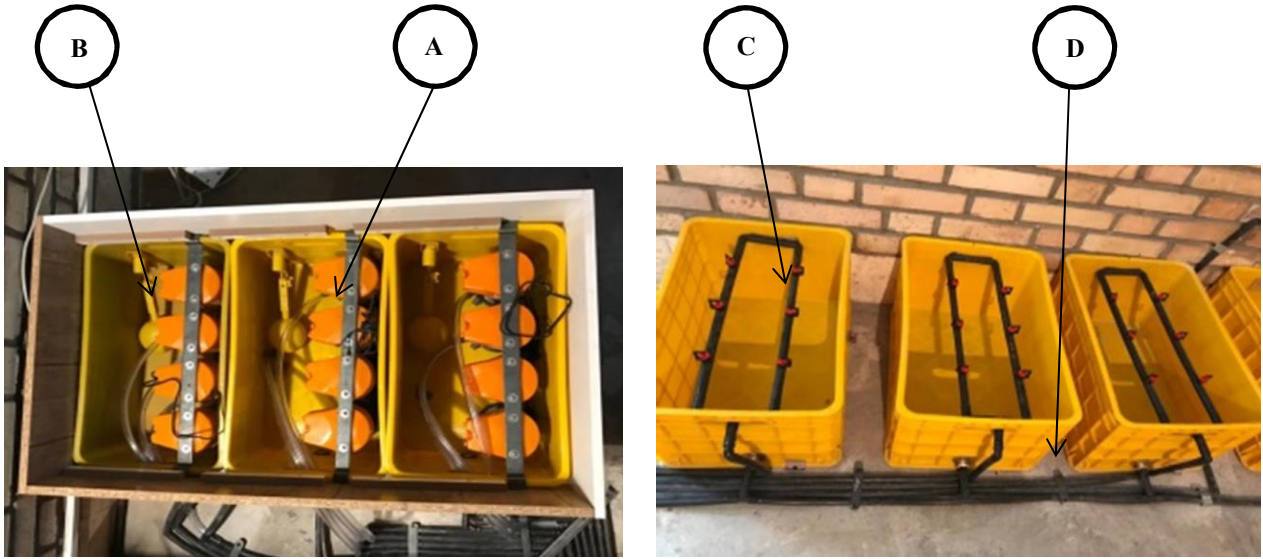

**Figure 3. Implementation of intelligent irrigation system (A) water transfer pumps, (B) water level control indicators, (C) centrifugal irrigation nozzles, (D) water transfer pipes**

## 2.3  Method of data collection and Internet of thing in the greenhouse

The data that needs to be measured in this greenhouse fall into two different categories. The first category is data on environmental conditions of plant growth, which are measured daily and automatically sent to the researcher to control the conditions. Measuring this data does not require the presence of the user. The second category is image data that is collected based on daily and weekly observations. Collecting this data requires the presence of the user in the greenhouse in compliance with health protocols.

Environmental conditions such as temperature, ambient humidity, light spectrum and intensity, water flow in cultivation boxes, and height and volume of water in water charging tanks are collecting by sensors such as SHT, GY-302, TCS3200, YF-S201, SRF05, as shown in Figure 4 (First category data).

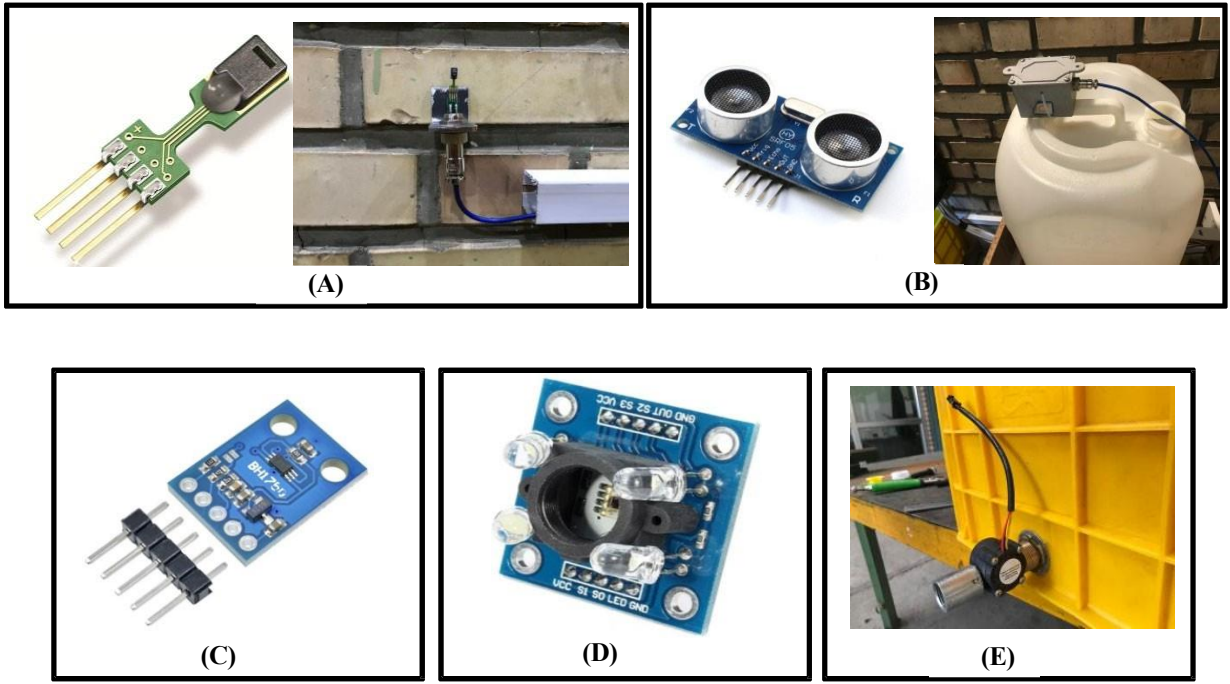

**Figure 4. Sensors for monitoring environmental conditions and irrigation in the greenhouse environment (A) SHT75 temperature and humidity monitoring sensor, and installation of SHT sensor in greenhouse environment (B) SRF05 water level detection sensor, and installation of protection box of SRF05 sensor on the top of nutrient solution storage tanks (C) GY-302 Light Intensity Monitoring Sensor (D) TCS3200 light spectrum monitoring sensor (E) Installation of YF-S201 water flow sensor**

This data is received by sensors and sent to the central processing unit. All information is sent to the cloud by a network module and stored on the memory card simultaneously.

### 2.4 Smart greenhouse actuators

In order to adjust and maintain the temperature and humidity of the greenhouse environment, the data received from the temperature and humidity sensors are processed, and according to that, the central board makes the decision on turning on and off the heating elements and the greenhouse fan, Arduino pro mini boards, and on/off relay switching modules. Also, the data received from SRF05 sensors and flow meter is sent to the user instantly through the Internet Shield to decide from the water level in the tanks to charge them quickly. The overview of the central processing unit is as shown in Figure 5.

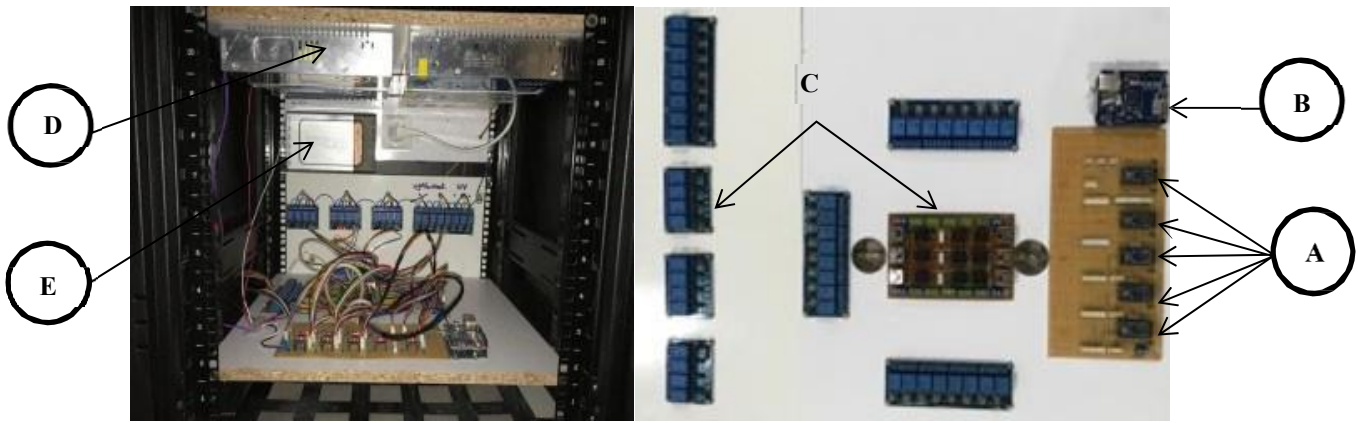

**Figure 5. Central control and processing unit of greenhouse (A) Arduino pro mini boards (B) Ethernet shield (C) Switching on and off relay modules (D) Power supplies (E) Internet router**

### 2.5 Sampling method of geranium plant and the role of transfer learning in plant leaf stress diagnosis

To ensure that the disease did not enter the greenhouse, cuttings with 4 or 6 healthy leaves of the geranium plant and free of any biotic and abiotic stress were separated from the mother plant, and the ornamental plants were transferred into the greenhouse and placed in the cultivation site. In each culture box, 12 cuttings of the geranium plant were placed, so 144 cuttings were placed in 9 boxes, and data collection began.

During non-destructive plant growth stages, imagery and sampling were started by an RGB camera 20 days after transferring the geranium plant cuttings to the greenhouse environment. Out of 144 cuttlefish cuttings, 133 cuttings passed the rooting and growth stages well. Images were taken of each plant at 4-day intervals until the 40th day. Since in the first days of the experiment, the number of healthy leaves was far more than two other classes, and in the last days of the experiment, the number of drought stress and rust leaves were far more than healthy leaves, the images captured every four days two have a homogeneous dataset of all leaves at the last of experiment in the last day Thus, 798 images were provided to evaluate the performance of the algorithm.

In addition, the images captured by the RGB camera are stored on the memory card and serve as input to the machine learning algorithm alongside other information received from the sensors (second category data).

According to the study results (Aversano et al., 2020), VGG-19 and Xception algorithms had the highest accuracy in detecting diseased leaves of the tomato plant. Based on this result, the VGG-19 algorithm was considered to distinguish healthy leaves from biotic and abiotic geranium stress leaves due to its high accuracy and speed of training. In addition to the robust VGG-19 algorithm, two powerful InceptionV3 and InceptionResNetV2 algorithms were used to compare their accuracy with each other. In order to identify the diseases in the network, a database consisting of 1000 images taken from geranium leaves in 5 industrial greenhouses was prepared. According to Figure 6, these images were divided into three categories: healthy leaves, drought stress leaves, and rust leaves in consultation with experts.

**(A)**   **(B)**   **(C)**

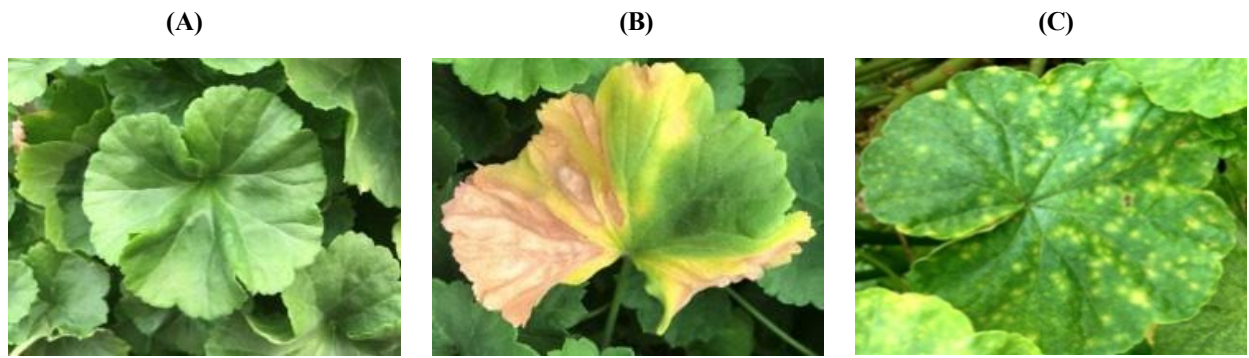

**Figure 6. Geranium leaves images taken from industrial greenhouses including (A) Healthy, (B) Drought stress, (C) Rust leaves**

As the VGG-19 algorithm's ability to classify data increases in proportion to the increase in the number of input data, the data number was increased to 5,000 images by rotation, zooming, horizontal flip, and vertical flip using the data augmentation technique. The network was formed by allocating 75% of the images to train, 15% to validation, and the remaining 10% to test the algorithm.

The exact obtained models were used to identify the three categories in the experimental greenhouse. The model prediction results were adapted to the expert diagnosis, and based on that, the accuracy of the proposed algorithm was calculated.

In this research, all environmental conditions of plant growth are reported instantly and remotely to the user by the IoT system. The images captured from the camera are processed by convolutional neural network algorithms and quickly report the symptoms of the disease in the early stages of plant growth. Finally, once informed of the disease, horticultural researchers will be able to report the appropriate concentration of the pesticide to the greenhouse owner to use and prevent the spread of the disease.

## 3. Result and discussion

### 3.1 Real-time monitoring with the internet of things system

In the greenhouse environment, the environmental conditions of plant growth are monitored instantly by the central processing unit, and it is possible to change the conditions instantaneously. It is also possible for the user to view parameters such as temperature, humidity, water flow in cultivation boxes, and water volume in nutrient solution charging tanks via the internet remotely. To achieve this goal, all data, after being collected by Arduino boards, is sent to the cloud via the Internet Shield and can be viewed online on the Ubidots platform, as shown in Figure 7.

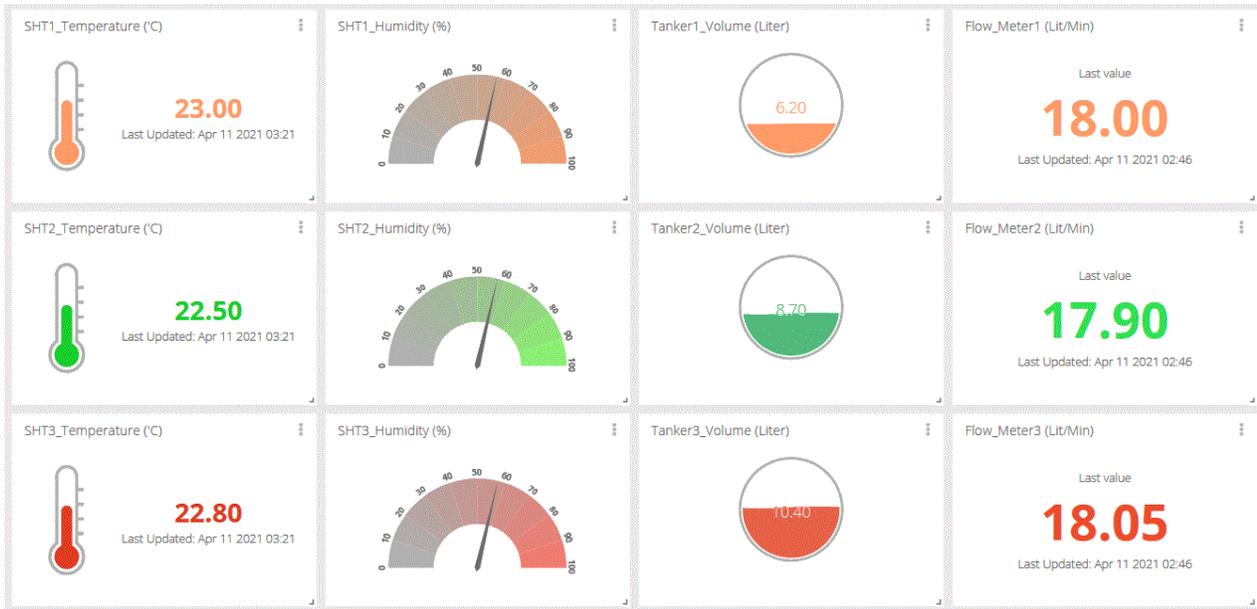

**Figure 7. Instantaneous monitoring of information received from sensors including temperature, humidity, water flow, the volume of water in tanks by the Ubidots platform**

### 3.2 VGG19, InceptionResNetV2, and InceptionV3 accuracy on Industrial greenhouses and experimental greenhouse dataset

Using the results obtained from industrial greenhouses, all three pre-trained algorithms VGG19, InceptionResNetV2, InceptionV3, were modified and retrained with the obtained data. The InceptionV3 algorithm achieved an accuracy of 0.7714 after completing 100 epochs, as shown in Figure 8.A.

In the next step, all the steps were repeated by replacing the InceptionResNetV2 algorithm. After passing 80 epochs, the accuracy obtained from the training and validation data converged towards each other, and according to Figure 8.B, finally achieved an accuracy of 0.8160.

In the final step, the VGG19 pre-trained algorithm recommended by (Aversano et al., 2020) was used to diagnose disease leaves. As shown in Figure 8.C, the algorithm was able to detect healthy leaves, drought stress, and rust leaves with 0.9294 accuracy after passing 100 epochs.

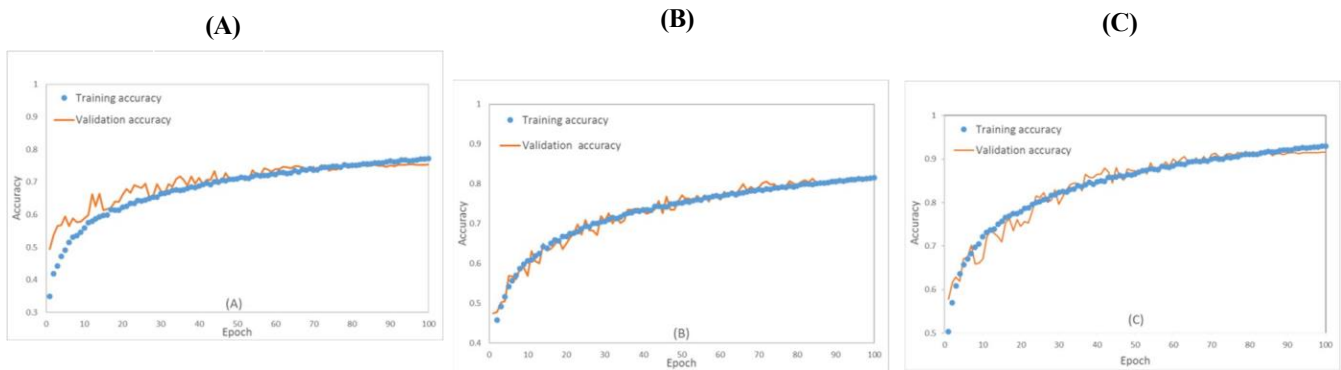

**Figure 8. Accuracy of (A) InceptionV3, (B) InceptionResNetV2, (C) VGG19, in training and validation phases on industrial greenhouse data**

Convolutional neural networks built on industrial greenhouse data were evaluated in the experimental greenhouse. In the experimental greenhouse, it was deliberately tried to have approximately equal shares of healthy, drought stress, and rust leaves to test the algorithm appropriately.

The InceptionV3 algorithm was tested on 798 images received from an experimental greenhouse and had the highest accuracy in detecting healthy leaves, 84.50%, and the lowest accuracy in detecting leaves under drought stress, 72.68%, and the final total accuracy on test data has been 78.44%. The confusion matrix of this algorithm can be seen in the test images in Figure 9.A.

Also, the InceptionResNetV2 algorithm had the highest accuracy in detecting healthy leaves, 86.84%, and the lowest accuracy in detecting leaves under drought stress, 72.68%, in the experimental greenhouse; and In addition, the total accuracy of this algorithm on test data was 81.07%. The confusion matrix of this algorithm can also be seen in Figure 9.B.

Finally, the VGG19 performance algorithm, which is the best choice based on previous research in this field, was evaluated on the data extracted from the greenhouse. The highest performance of this algorithm in the diagnosis of healthy leaves was 94.44%, and the lowest performance in the diagnosis of leaves under drought stress was 75.60%, and its overall performance has been 86.34%. The confusion matrix of this algorithm and its performance in classifying leaves from each other can be seen in Figure 9.C.

**(A)**            **(B)**            **(C)**

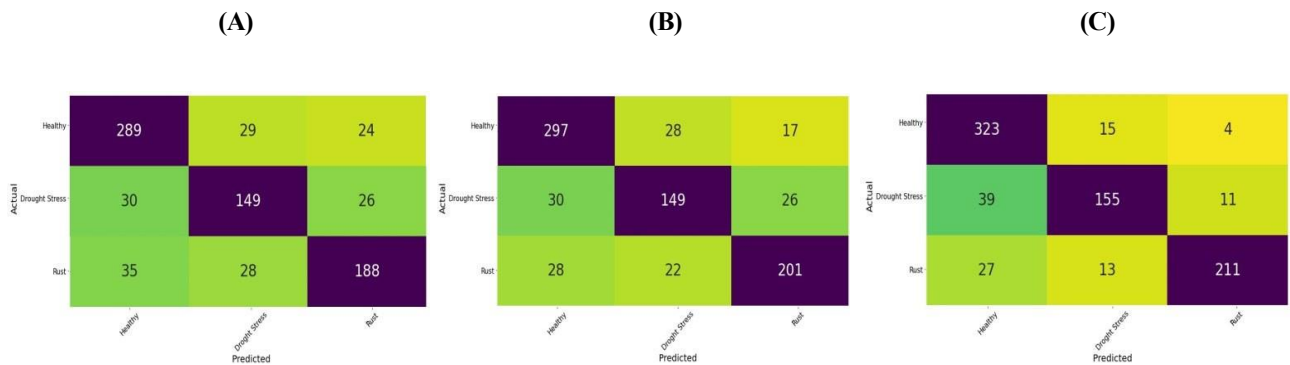

**Figure 9. (A) InceptionV3, (B) InceptionResNetV2, and (C) VGG-19 confusion matrices on experimental greenhouse test data**

Figure 10 shows the share of the number of leaves belonging to each class of healthy leaves, drought stress leaves, and rust leaves in percent after the end of the experiment in the experimental greenhouse. Finally, as shown in Figure 10, it is clear that all three algorithms have the highest accuracy in detecting healthy leaves and the lowest accuracy in detecting drought stress leaves, one of the main reasons for which may be more healthy leaves from the beginning to the end of the experiment in the experimental greenhouse. Also, as expected, the VGG19 algorithm had the highest detection for all leaves classes, and the Inception algorithm had the lowest accuracy.

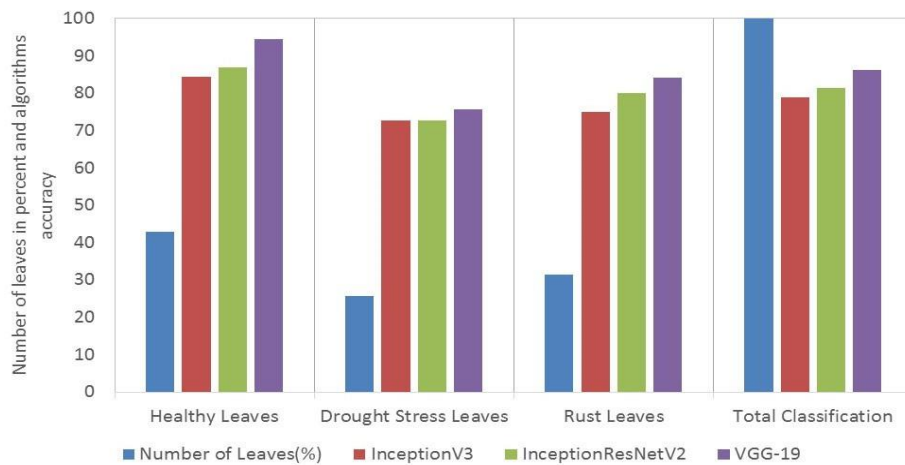

**Figure 10. The number of leaves in each class in percent, and the accuracy of VGG19, InceptionResNetV2, InceptionV3 algorithms in detecting healthy, drought stress leaves and rust leaves and their total classification accuracy in the experimental greenhouse**

# 4. Conclusion

One of the main problems of greenhouses is the presence of pandemic diseases due to the constant presence of workers in them, and generally, there is no way to diagnose diseases early. Also, if it is possible to have an accurate timing of irrigation in the aeroponic system and all environmental conditions of plant growth are reported instantly, the plant will definitely reach its optimal growth.

In this research, an intelligent irrigation system was installed, with the help of which irrigation boxes are irrigated according to the user-defined time discretion with a specific schedule. Also, all environmental conditions of plant growth are reported online to the user to make a timely decision as soon as a sudden parameter change.

Algorithms VGG-19, InceptionResNetV2, and InceptionV3 were also considered to diagnose geranium disease and stress in the greenhouse environment. Imaging 5 industrial greenhouses and the classification of greenhouse leaves into three healthy leaves, drought stress leaves, and rust leaves classes was performed. After training the algorithms, VGG19 achieved the highest accuracy, 0.92, and InceptionResNetV2 achieved the accuracy of 0.81, and Inception achieved the lowest accuracy, 0.77. In all algorithms, healthy leaves were classified with the highest accuracy because healthy leaves had the highest number in all tests.

In the experimental greenhouse, trained algorithms were used to predict disease in the greenhouse environment. The results of the confusion matrices showed that the VGG19 algorithm had the highest accuracy, 86.34%, and the InceptionV3 algorithm had the lowest accuracy, 78.44%, on the experimental greenhouse test data.

However, the detection accuracy of the VGG-19 algorithm is lower than expected. It is expected that if the structure of the convolutional neural networks is reviewed with images of plants grown in the greenhouse and the number of images obtained increases, its accuracy in diagnosing the type of disease will improve significantly.

In the future, in order to increase the accuracy of algorithms and diagnosis of the disease in earlier stages, the algorithm will be trained on more data. Also, to increase work flexibility, more varieties and diseases of geranium plants will be examined. In addition to increasing the number of varieties and geranium diseases, using other plants in the greenhouse environment to prepare a database of common plant diseases is also recommended.